\documentclass[letterpaper, 10 pt, conference]{class/ieeeconf}  

\IEEEoverridecommandlockouts                              
\usepackage{siunitx}  
\usepackage{graphicx}
 \graphicspath{./figures/robot_demonstration}
\usepackage{caption}
\usepackage{amsmath}
\usepackage{amssymb}
\usepackage{latexsym}
\usepackage{url}
\usepackage{cite}
\usepackage{relsize}
\usepackage{multirow}
\usepackage{afterpage}
\usepackage{ifthen}
\usepackage{graphicx}
\usepackage{algpseudocode,algorithm,algorithmicx}
\usepackage{subfigure}
\usepackage{flushend}
\usepackage{epstopdf}
\usepackage{stackengine}
\usepackage{textcomp} 

\usepackage{amsmath}
\usepackage{tcolorbox}
\usepackage{enumerate}
\let\labelindent\relax
\usepackage[shortlabels]{enumitem}\usepackage{amssymb}
\usepackage{latexsym}
\usepackage{url}
\usepackage{cite}
\usepackage{relsize}
\usepackage{multirow}
\usepackage{afterpage}
\usepackage{ifthen}
\usepackage{graphicx}
\usepackage{algpseudocode,algorithm,algorithmicx}
\usepackage{subfigure}

\usepackage{epstopdf}
\usepackage{stackengine}
\usepackage{textcomp} 
\usepackage{pifont}
\newcommand{\cmark}{\ding{51}}%
\newcommand{\xmark}{\ding{55}}%
\usepackage{threeparttable,booktabs}
\usepackage{gensymb}
\usepackage{booktabs}
\usepackage{makecell}
\usepackage{hhline}
\usepackage{courier}
\usepackage{lipsum}
\usepackage{diagbox}
\usepackage{xcolor} 
\usepackage{xspace}

\usepackage[dvipsnames]{xcolor}

\usepackage{gensymb}
\usepackage{booktabs}
\usepackage{makecell}
\usepackage{hhline}
\usepackage{courier}
\usepackage{lipsum}
\usepackage{tcolorbox}
\newtcbox{\inlinecode}{on line, boxsep=1pt, arc=3pt, colback=black, colframe=black, coltext=white}

\makeatletter
\let\NAT@parse\undefined
\makeatother
\usepackage[bookmarks=false, linkcolor=blue, urlcolor=blue, citecolor=blue]{hyperref} 

\hypersetup{
    colorlinks=true,
    linkcolor=red,
    filecolor=magenta,      
    urlcolor=blue,
    pdfstartview={FitH},
    citecolor =blue
    }

\graphicspath{{./figures/exp}}

\usepackage{xspace}

\title{\LARGE \bf Beyond Visual Grasping: Benchmarking Complex Grasping from Detection to Execution} \vspace{-8ex}
\author{H. Zhang$^{1}$, K. Nguyen$^{2}$, C. Munasinghe$^{3}$, B. Hela$^{4}$, T. Li$^{1}$, Z. Luo$^{1}$, H. Nguyen$^{5}$,
H.W van de Venn$^{3}$,\\ Y. Zheng$^{1}$, R. Prakash$^{4}$, T. D. Ta$^{6,7}$, A. Nguyen$^{1}$, B. Huang$^{1}$ 
\thanks{$^{1}$University of Liverpool, UK}
\thanks{$^{2}$Mohamed bin Zayed University of Artificial Intelligence, UAE}
\thanks{$^{3}$Zurich University of Applied Science, Switzerland}
\thanks{$^{4}$Indian Institute of Science, India}
\thanks{$^{5}$University of Information Technology, Vietnam}
\thanks{$^{6}$The University of Tokyo, Japan}
}

\begin{document}

\newtheorem{problem}{Problem}
\newtheorem{lemma}{Lemma}
\newtheorem{theorem}[lemma]{Theorem}
\newtheorem{claim}{Claim}
\newtheorem{corollary}[lemma]{Corollary}
\newtheorem{definition}[lemma]{Definition}
\newtheorem{proposition}[lemma]{Proposition}
\newtheorem{remark}[lemma]{Remark}
\newenvironment{LabeledProof}[1]{\noindent{\it Proof of #1: }}{\qed}

\def\beq#1\eeq{\begin{equation}#1\end{equation}}
\def\bea#1\eea{\begin{align}#1\end{align}}
\def\beg#1\eeg{\begin{gather}#1\end{gather}}
\def\beqs#1\eeqs{\begin{equation*}#1\end{equation*}}
\def\beas#1\eeas{\begin{align*}#1\end{align*}}
\def\begs#1\eegs{\begin{gather*}#1\end{gather*}}

\newcommand{\poly}{\mathrm{poly}}
\newcommand{\eps}{\epsilon}
\newcommand{\e}{\epsilon}
\newcommand{\polylog}{\mathrm{polylog}}
\newcommand{\rob}[1]{\left( #1 \right)} 
\newcommand{\sqb}[1]{\left[ #1 \right]} 
\newcommand{\cub}[1]{\left\{ #1 \right\} } 
\newcommand{\rb}[1]{\left( #1 \right)} 
\newcommand{\abs}[1]{\left| #1 \right|} 
\newcommand{\zo}{\{0, 1\}}
\newcommand{\zonzo}{\zo^n \to \zo}
\newcommand{\zokzo}{\zo^k \to \zo}
\newcommand{\zot}{\{0,1,2\}}
\newcommand{\en}[1]{\marginpar{\textbf{#1}}}
\newcommand{\efn}[1]{\footnote{\textbf{#1}}}
\newcommand{\vecbm}[1]{\boldmath{#1}} 
\newcommand{\uvec}[1]{\hat{\vec{#1}}}
\newcommand{\thv}{\vecbm{\theta}}
\newcommand{\junk}[1]{}
\newcommand{\var}{\mathop{\mathrm{var}}}
\newcommand{\rank}{\mathop{\mathrm{rank}}}
\newcommand{\diag}{\mathop{\mathrm{diag}}}
\newcommand{\tr}{\mathop{\mathrm{tr}}}
\newcommand{\acos}{\mathop{\mathrm{acos}}}
\newcommand{\atantwo}{\mathop{\mathrm{atan2}}}
\newcommand{\SVD}{\mathop{\mathrm{SVD}}}
\newcommand{\quadf}{\mathop{\mathrm{q}}}
\newcommand{\linterp}{\mathop{\mathrm{l}}}
\newcommand{\sgn}{\mathop{\mathrm{sign}}}
\newcommand{\sym}{\mathop{\mathrm{sym}}}
\newcommand{\avg}{\mathop{\mathrm{avg}}}
\newcommand{\mean}{\mathop{\mathrm{mean}}}
\newcommand{\erf}{\mathop{\mathrm{erf}}}
\newcommand{\grad}{\nabla}
\newcommand{\R}{\mathbb{R}}
\newcommand{\defeq}{\triangleq}
\newcommand{\dims}[2]{[#1\!\times\!#2]}
\newcommand{\sdims}[2]{\mathsmaller{#1\!\times\!#2}}
\newcommand{\udims}[3]{#1}
\newcommand{\udimst}[4]{#1}
\newcommand{\com}[1]{\rhd\text{\emph{#1}}}
\newcommand{\ind}{\hspace{1em}}
\newcommand{\argmin}[1]{\underset{#1}{\operatorname{argmin}}}
\newcommand{\floor}[1]{\left\lfloor{#1}\right\rfloor}
\newcommand{\step}[1]{\vspace{0.5em}\noindent{#1}}
\newcommand{\quat}[1]{\ensuremath{\mathring{\mathbf{#1}}}}
\newcommand{\norm}[1]{\left\lVert#1\right\rVert}
\newcommand{\ignore}[1]{}
\newcommand{\specialcell}[2][c]{\begin{tabular}[#1]{@{}c@{}}#2\end{tabular}}
\newcommand*\Let[2]{\State #1 $\gets$ #2}
\newcommand{\algorithmicbreak}{\textbf{break}}
\newcommand{\Break}{\State \algorithmicbreak}
\newcommand{\ra}[1]{\renewcommand{\arraystretch}{#1}}

\renewcommand{\vec}[1]{\mathbf{#1}} 

\algdef{S}[FOR]{ForEach}[1]{\algorithmicforeach\ #1\ \algorithmicdo}
\algnewcommand\algorithmicforeach{\textbf{for each}}
\algrenewcommand\algorithmicrequire{\textbf{Require:}}
\algrenewcommand\algorithmicensure{\textbf{Ensure:}}
\algnewcommand\algorithmicinput{\textbf{Input:}}
\algnewcommand\INPUT{\item[\algorithmicinput]}
\algnewcommand\algorithmicoutput{\textbf{Output:}}
\algnewcommand\OUTPUT{\item[\algorithmicoutput]}


\twocolumn[
{%
\renewcommand\twocolumn[1][]{#1}%
\maketitle

\begin{center}
    \centering
    \vspace{-5ex}
\includegraphics[width=0.99\textwidth,height=0.58\linewidth]{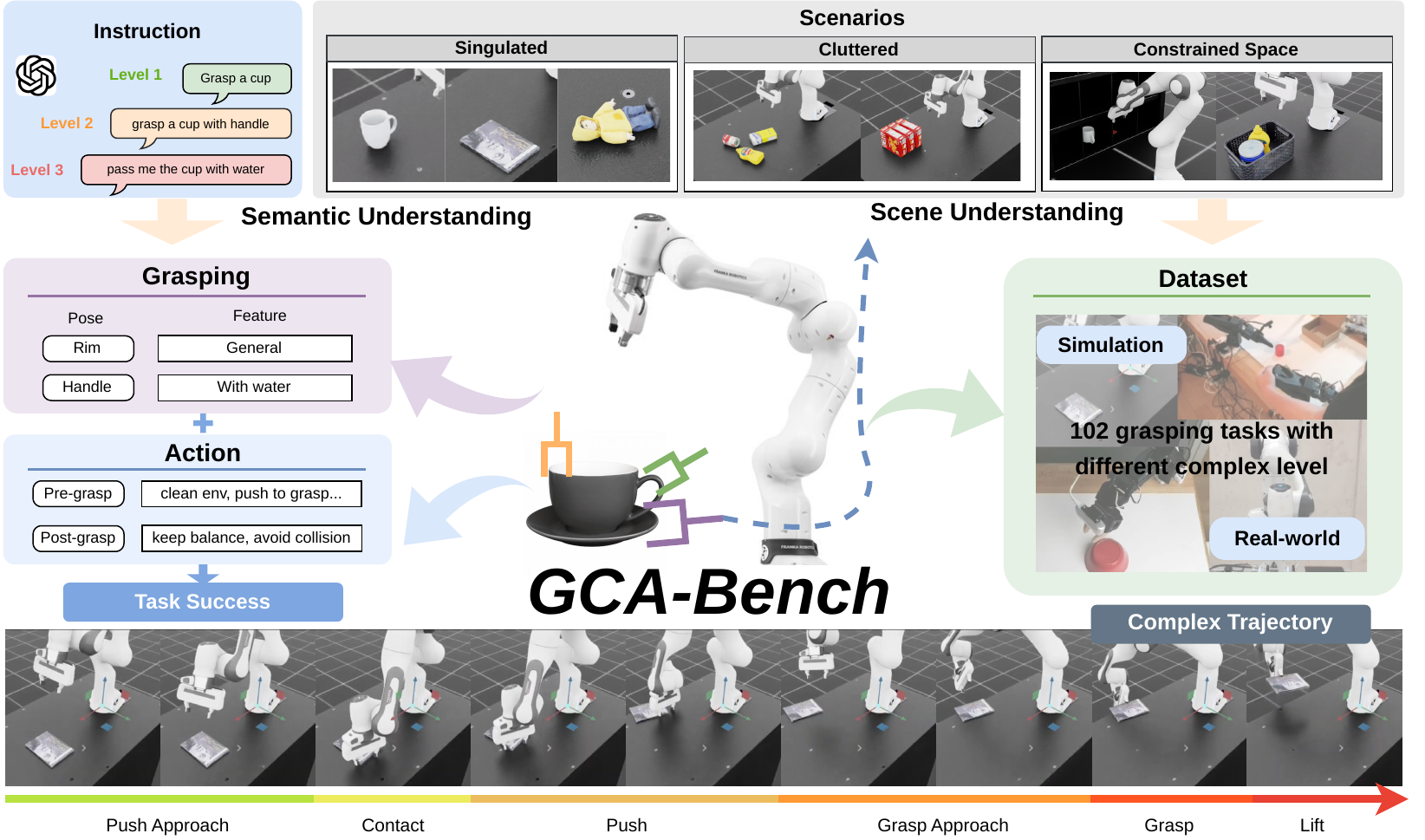}
    \vspace{2ex}  
    \captionof{figure}{We propose GCA-Bench, a benchmark for evaluating complex robotic grasping skills as a multi-stage process.} 
    \label{fig:IntroVis}
\end{center}
}]
\renewcommand{\thefootnote}{\arabic{footnote}}
\footnotetext[1]{University of Liverpool, UK}
\footnotetext[2]{Mohamed bin Zayed University of Artificial Intelligence, UAE}
\footnotetext[3]{Zurich University of Applied Science, Switzerland}
\footnotetext[4]{Indian Institute of Science, India}
\footnotetext[5]{University of Information Technology, HCMC, Vietnam}
\footnotetext[6]{Keio University, Japan}
\let\thefootnote\relax\footnotetext{AN and TT were supported by the Royal Society ISPF International Collaboration Awards (ICA\textbackslash R1\textbackslash 231067)}
\begin{abstract} 
Robust robotic grasping remains a fundamental challenge for complex real-world applications. Recent advances in large-scale models demonstrate promising capabilities for reasoning in robotic tasks. However, existing benchmarks for grasping primarily focus on isolated, visual-based grasp pose detection, failing to capture the complexity of grasping tasks that require multi-step reasoning and semantic understanding during execution. To address this gap, we propose GCA-Bench, a benchmark featuring challenging \textit{grasping with complex action} scenarios that involve both scene-level reasoning and semantic constraints. GCA-Bench enables the evaluation of recent large foundation models under the same settings. To demonstrate the effectiveness of our new benchmark, we implement a diverse set of baselines, ranging from traditional grasp detection pipelines to end-to-end learning methods. Empirical studies achieve success rates below 70\% on complex grasping scenarios, underscoring critical limitations. In addition, we propose new evaluation metrics, analyze critical failure models, and provide insights to guide the development of more robust and generalizable grasping strategies. Our project is available at \href{https://airvlab.github.io/GCA-Bench/}{https://airvlab.github.io/GCA-Bench/}

\end{abstract}


\section{INTRODUCTION} \label{Sec: intro}
 \vspace{-1ex}
 Grasping constitutes a core capability for autonomous robots with several real-world applications~\cite{xie2023learning,dong2023review}. Recent advances in robotic grasping span multiple directions, including learning-based grasp pose detection and the adoption of foundation models for semantic task understanding~\cite{du2021vision}. These developments have substantially improved grasp success in standard benchmarks. Despite these advances, grasping performance remains fragile in complex scenarios~\cite{han2024fetchbench}, including manipulating objects with difficult geometries such as flat cards and thin sheets, operating in cluttered environments, retrieving items from confined spaces, or satisfying language-specified task constraints~\cite{newbury2023deep,efendi2025technological}.

\begin{table*}[h]
\centering
\caption{Comparison of grasping benchmarks and datasets.}
\vspace{1em}
\resizebox{0.95\textwidth}{!}{%
\begin{tabular}{l  l c c c c c c c }
\toprule
\textbf{Benchmark} & \textbf{Focus} &  \textbf{Language\textsuperscript{1}}  & \textbf{Trajectory\textsuperscript{2}} & \textbf{Grasp Evaluation\textsuperscript{3}}& \textbf{Complex Grasping\textsuperscript{4}} & \textbf{Sim Data\textsuperscript{5}} & \textbf{Real Data\textsuperscript{6}}  \\ \midrule
RLBench~\cite{james2020rlbench}  & Manipulation          & \xmark &\checkmark & \xmark & \xmark & \checkmark & \xmark \\ 
Robocasa~\cite{nasiriany2024robocasa} & Manipulation          & \xmark & \checkmark & \xmark & \xmark & \checkmark & \xmark \\ 
LIBERO~\cite{liu2023libero} & Manipulation           & \checkmark   & \checkmark & \xmark & \xmark & \checkmark & \xmark \\ 
VLAbench~\cite{zhang2024vlabench} & Generality            &\checkmark  & \checkmark & \xmark & \xmark & \checkmark & \xmark \\ 
COLOSSEUM~\cite{pumacay2024colosseum} & Perturbation        & \xmark & \checkmark & \xmark & \xmark & \checkmark & \xmark \\\midrule
GraspNet-1Billion~\cite{fang2020graspnet} & Grasp pose detection & \xmark& \xmark & \checkmark & \xmark & \xmark & \checkmark \\ 
GraspAnything ~\cite{vuong2024grasp}& Grasp pose detection    &\checkmark& \xmark & \checkmark & \xmark & \checkmark & \xmark \\ 
FMB~\cite{luo2025fmb} & Assembly problems     & \xmark& \checkmark & \checkmark & \checkmark & \xmark & \checkmark \\ 
Fetchbench~\cite{han2024fetchbench} & Fetching          & \xmark& \checkmark & \checkmark & \xmark & \xmark & \xmark \\ 
SynGrasp-1B~\cite{deng2025graspvla} & Grasp pose detection             & \xmark& \checkmark & \checkmark & \xmark & \checkmark & \xmark \\ \midrule
\textbf{GCA-Bench} (ours) & Complex grasping      &\textcolor{orange}{\cmark}& \textcolor{orange}{\cmark} & \textcolor{orange}{\cmark} & \textcolor{orange}{\cmark} &\textcolor{orange}{\cmark} & \textcolor{orange}{\cmark} \\ \bottomrule
\end{tabular}
}

\textsuperscript{1} Language instructions.\
\textsuperscript{2} Task execution trajectory is considered.\
\textsuperscript{3} Grasp ability evaluation.\
\textsuperscript{4} Complexity categorization of grasping tasks.\
\textsuperscript{5} Synthetic or simulated data collection.\
\textsuperscript{6} Real-world data collection.\
\label{tab:benchmark_comparison}
\end{table*}

We define complex grasping as grasping that requires scene and semantic reasoning capabilities beyond isolated grasp pose prediction to overcome geometric, spatial, or semantic difficulty. 
Existing efforts to address these challenges generally follow two complementary paradigms. One emphasizes hardware specialization, introducing suction-based~\cite{cao2021suctionnet}, multi-fingered~\cite{lundell2021multi}, or adaptive end-effectors~\cite{xu2021adagrasp} to expand physical capabilities. The other seeks to enhance algorithmic intelligence for parallel-jaw grippers, which continue to dominate practical robotic deployments. Algorithmic strategies include data-driven grasp pose prediction integrated with motion planning~\cite{fang2020graspnet, vuong2024grasp, depierre2018jacquard}, leveraging large-scale foundation models for semantic reasoning~\cite{vuong2024language, ngyen2023open, stone2023open}, and end-to-end architectures that unify perception and control~\cite{mohammed2020review,hua2021learning}. 
While these techniques show promise, it remains insufficiently understood how these approaches perform across diverse and complex grasping scenarios. This gap motivates the need for systematic evaluation frameworks capable of exposing failure cases and informing the next generation of algorithmic designs for grasping systems.

Current benchmarking efforts in robotic grasping remain largely focused on visual grasp pose detection~\cite{vuong2024grasp,liu2021ocrtoc,nguyen2024lightweight}, where success is evaluated in isolation rather than across the full manipulation sequence. Recent research~\cite{han2024fetchbench} has demonstrated that even state-of-the-art methods exhibit substantial performance degradation when evaluated across the entire grasping pipeline. Furthermore, recent large models such as $\pi_0$~\cite{black2024pi_0}, which bypass explicit grasp detection in favor of direct action prediction, are making traditional detection-oriented benchmarks increasingly inadequate. 
This gap underscores the need for a unified benchmarking framework that enables fair and comprehensive evaluation of modern grasping algorithms.
To address this gap, we introduce GCA-Bench, a benchmark designed to evaluate grasping with complex actions across multiple levels of difficulty. Unlike conventional benchmarks that focus solely on grasp pose detection, GCA-Bench requires integrated reasoning over trajectory planning, including spatial reasoning and multi-step coordination, as well as semantic task understanding. 
Our contributions include: 

\begin{itemize}
    \item We introduce GCA-Bench, a complex benchmark to evaluate robotic grasping with complex action scenarios, encompassing both scene-level reasoning and semantic constraints throughout the full execution pipeline.
    \item We conduct intensive experiments and analysis to reveal critical limitations of existing methods and provide insights to develop more robust and generalizable robotic grasping systems for real-world applications.
\end{itemize}





\section{Related Work} \label{Sec: related_work}
\textbf{Complex Grasping.}
Conventional grasping systems separate the problem into grasp pose detection~\cite{vuong2024language,fang2023anygrasp,nguyen2024language} and collision-free motion planning~\cite{han2024fetchbench, sundaralingam2023curobo}. While effective in detection benchmarks, such systems assume static execution and lack online adaptation or holistic reasoning over the entire manipulation process~\cite{geng2023sage, liu2024efficient}. In more complex settings, grasping must handle occlusions, clutter, ungraspable features, and spatial constraints, while aligning grasp strategies with task intent~\cite{han2024fetchbench, efendi2025technological, li2024semgrasp}. End-to-end policy learning attempts to close this gap by mapping perception and language inputs directly to actions~\cite{gao2025end}. Reinforcement learning enables adaptive strategies such as push–grasp sequences and recovery behaviors~\cite{lobbezoo2023simulated, sekkat2024review, huang2024novel}, while imitation learning improves data-efficient acquisition of contact-rich skills~\cite{wang2024robot, song2020grasping}. More recently, foundation models have enhanced semantic grounding and generalization for language-conditioned grasping~\cite{deng2025graspvla, zhong2025dexgraspvla, murali2025graspgen, zheng2024gaussiangrasper}. Nevertheless, robust adaptation under strict geometric and semantic constraints remains an open challenge~\cite{han2024fetchbench, efendi2025technological}.

\textbf{Grasping Benchmarks.}
Over the years, many benchmarking frameworks have been developed for manipulation and grasping tasks. As shown in Table~\ref{tab:benchmark_comparison}, manipulation benchmarking works such as LIBERO~\cite{liu2023libero} and VLABench~\cite{zhang2024vlabench} provide comprehensive task-level evaluation but focus primarily on high-level manipulation policies and long-horizon tasks rather than low-level grasping execution. However, current benchmarks on grasping, such as GraspNet-1Billion~\cite{fang2020graspnet} and Grasp-anything~\cite{vuong2024grasp}, mainly focus on grasp pose detection without considering the execution. FetchBench~\cite{han2024fetchbench} represents a notable advance in addressing the integration of grasping and motion planning within fetching tasks. However, it only focused on the difficulty of collision avoidance rather than on the complexity of the tasks themselves. In contrast,  GCA-Bench is the first benchmark designed for complex grasping tasks from detection to execution.

\begin{figure*}[t]
    \centering
    \includegraphics[width=0.92\linewidth]{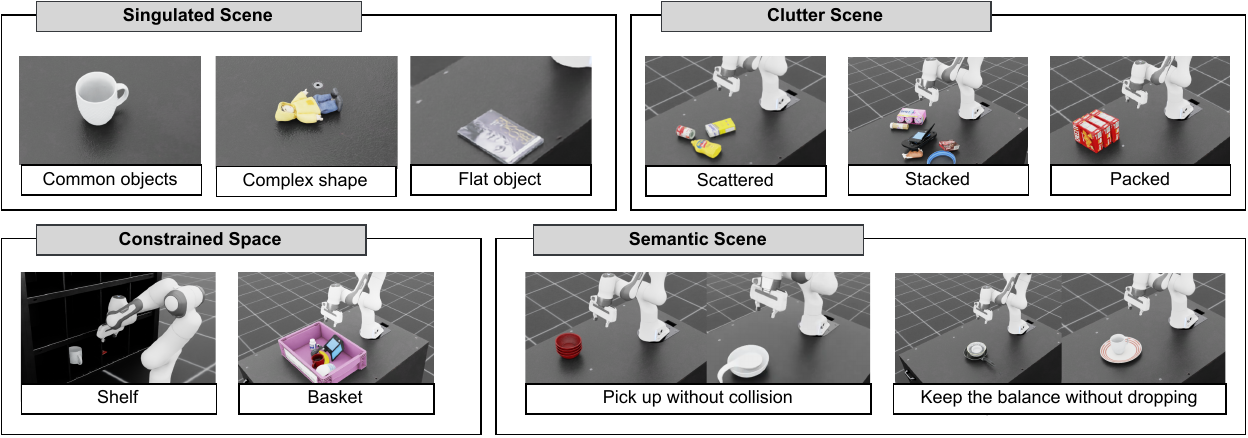}
    \vspace{2ex}
    \caption{Overview of scenario design in GCA-Bench. The benchmark comprises four categories of manipulation tasks: (\textit{i}) \textit{Singulated Scene}, involving isolated objects of varying shapes and poses; (\textit{ii}) \textit{Cluttered Scene}, where objects are scattered, stacked, or densely packed; (\textit{iii}) \textit{Constrained Space}, requiring precise manipulation in constrained environments such as shelves or baskets; and (\textit{iv}) \textit{Semantic Scene}, requiring the robot to strictly follow the language instructions.
    }
    \label{fig:task design}
    \vspace{-1ex}
\end{figure*}

We propose GCA-Bench, a robotic benchmark for challenging grasping scenarios that require understanding how humans perform the task. Unlike benchmarks such as LIBERO~\cite{liu2023libero}, which mainly target common everyday manipulation skills (e.g., opening a microwave), GCA-Bench emphasizes human-like task understanding for difficult and realistic real-world robotic grasps.

\section{The GCA-Bench} \label{Sec: method}

\subsection{Scenario Design}\label{task design}
To enable comprehensive evaluation of robotic grasping and manipulation, we design 102 task scenarios with varying levels of difficulty, ranging from singulated objects to cluttered environments, constrained narrow spaces, and semantic tasks requiring contextual reasoning (Fig.~\ref{fig:task design}) with 102 tasks in total.

\textbf{Singulated Scene.} This basic scenario involves grasping and lifting single objects placed on a flat surface. The objects are selected to span a variety of properties, such as common objects which are mostly from YCB datasets\cite{calli2015benchmarking}, and objects with complex shape such as some irregularly shaped toys and flat objects that are inherently challenging to grasp. To ensure robust evaluation, we vary object pose and placement. For example, a bowl may be positioned upright or upside down, significantly affecting grasp difficulty. 

\textbf{Cluttered Scene.} Cluttered environments simulate real-world scenarios where objects are overlapping or stacked. We use the same object set to construct clutter scenes with increasing complexity.
    These include:
    \begin{itemize}
        \item \textit{Scattered Scene}: Objects are randomly distributed across the workspace with sufficient spacing.
        \item \textit{Stacked Scene}: Objects are piled tightly, resulting in severe occlusions and increased risk of collision.
        \item \textit{Packed Scene}: Objects are densely arranged along axes, leaving minimal clearance for gripper insertion.
    \end{itemize}
    These settings challenge the robot to detect objects under occlusion and plan access paths.

\textbf{Constrained Space.} Constrained environments, such as shelves or baskets, test the robot's ability to operate within limited spatial bounds. These scenarios demand precise motion planning, adaptive collision avoidance, and safe navigation. The difficulty is determined by object placement and surrounding obstructions. For example, shelf-mounted objects may be occluded by frames or adjacent items, while baskets often contain tightly packed objects requiring careful maneuvering to avoid damage. Objects placed in corners or against walls further restrict feasible grasp trajectories.

\textbf{Semantic Scene.} We design challenging tasks specifically targeting semantic understanding. Semantic scenarios evaluate the robot’s ability to reason about task-level context beyond geometric constraints, requiring high-level decision making in both grasp pose selection and trajectory planning. For instance, in a cup set task, the robot must grasp without disturbing balance to prevent spillage, which demands timely feedback and online adjustment. Similarly, tasks such as retrieving a fragile object from a cluttered scene require precise motion planning and careful collision avoidance.
\begin{figure*}
    \centering
    \includegraphics[width=0.90\linewidth]{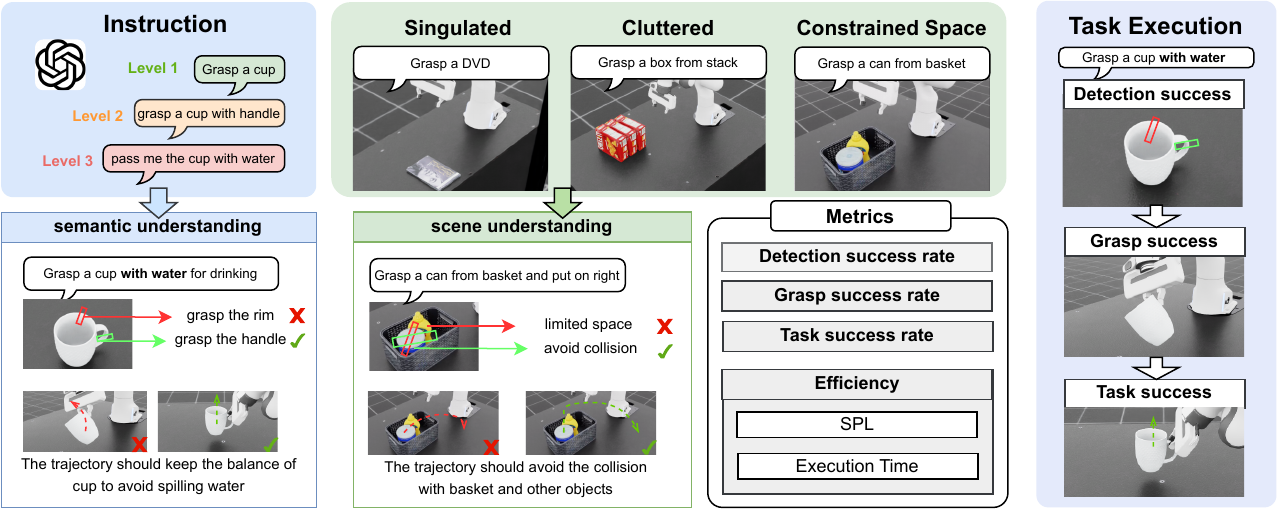}
    \vspace{2ex}
    \caption{Tasks are designed to evaluate both semantic and scene understanding. Execution is assessed along a pipeline of detection, grasping, and task-level success. Complex tasks require not only correct grasping but also proper trajectory execution, as illustrated by the examples. Performance is measured using detection success rate, grasp success rate, task success rate, and efficiency metrics including SPL and execution time.}
    \label{fig:evaluation}
    \vspace{-1ex}
\end{figure*}

\subsection{Instruction Generation}  
To enable language-driven human–robot interaction, we use a large language model (ChatGPT-4o mini~\cite{openai2021chatgpt}) to generate task instructions at three levels of complexity. This choice is motivated by how humans naturally communicate with robots: instructions are often expressed in flexible, high-level language, may omit details, and vary widely in specificity. ChatGPT-4o mini~\cite{openai2021chatgpt} is well-suited for this because it can produce diverse, fluent, and context-aware instructions that resemble natural user prompts, rather than templated command scripts. Generated instructions were manually reviewed to remove ambiguous object references, mismatched complexity levels, and ungrammatical phrasing.

\textbf{Level 1 - Basic Instructions.} Simple commands that specify the target object, e.g., “Grasp the cup.” These instructions require object recognition and basic grasp execution without further context.

\textbf{Level 2 – Specific Instructions with Constraints.} These instructions include explicit requirements, such as “Grasp the cup by the handle” or “Pick up the book from the top edge.” Robots must interpret object affordances and plan grasp strategies that satisfy constraints.

\textbf{Level 3 – Complex Task Instructions.} High-level semantic tasks require reasoning and multi-step planning. Examples include “Pass me a cup with water” or “Give me a knife.” These tasks demand context-aware grasping, e.g., avoiding tilting a full cup or grasping a knife safely by the handle, requiring semantic understanding of function, safety, and trajectory planning.

\subsection{Environment Setup}
\begin{figure}[h]
    \centering
    \includegraphics[width=0.94\linewidth]{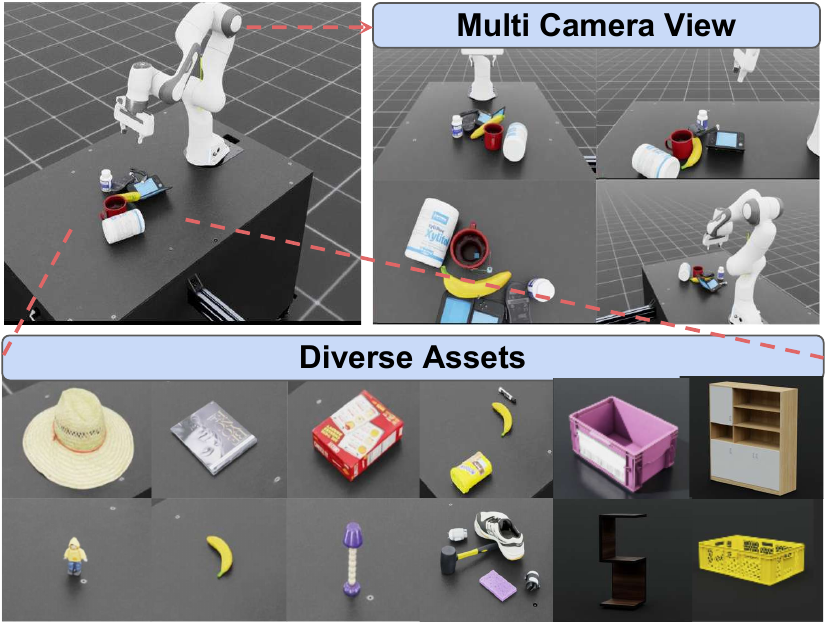}
    \vspace{3ex}
    \caption{Simulation setup in IsaacLab with four cameras: front, wrist-mounted, side, and over-the-shoulder. Task environments are constructed using diverse assets with varying physical properties from NVIDIA Omniverse.}
    \label{fig:simulation setup}
    \vspace{0ex}
\end{figure}

We utilize NVIDIA Isaac Lab~\cite{mittal2023orbit} as the primary simulation platform for robot learning and policy training. Isaac Lab is a GPU-accelerated, open-source framework designed to unify and simplify robotics research workflows. Our benchmark is designed to flexibly integrate different robotic arms. In this work, we focus on the 7-DOF Franka Emika Panda with a Franka gripper. To provide comprehensive visual coverage of the workspace and meet the requirements of different algorithms, the robot is set up with four RGB-D cameras as shown in Fig.~\ref{fig:simulation setup}: a wrist camera mounted on the gripper, a third-view camera positioned in front of the arm, a side-view camera, and an over-shoulder camera. The end-effector pose is represented using 3D coordinates for position and quaternions for orientation. For standardising the evaluation, we inherited annotated objects from the MultiGripperGrasp dataset~\cite{casas2024multigrippergrasp} with 30.4 million grasps over 345 objects using 11 different grippers. To construct diverse evaluation scenes, we additionally leverage various assets from Nvidia Omniverse, including containers and furniture such as tables, cabinets, and shelves.

\subsection{Evaluation Metrics}
As shown in Fig.~\ref{fig:evaluation}, our benchmark evaluates grasping methods across varying task difficulties, considering both semantic understanding and scene reasoning. Beyond grasping success, we also assess task execution, providing insights into the limitations of current grasping pipelines and the gap between perception and execution. To evaluate full grasp execution, we design evaluation metrics at several gradualities, from detection, grasping, to task achievement. Success is defined not only by grasp completion but also by grasp quality and subsequent actions, such as maintaining object balance or avoiding collisions. In this setting, the robot must execute grasps that satisfy semantic requirements rather than merely secure the object. Comparing performance across instruction levels allows us to isolate semantic reasoning from low-level manipulation, providing a more nuanced assessment of an algorithm’s ability to interpret and act on task semantics in real-world contexts.

\textbf{Metrics.} We evaluate performance using several complementary metrics. All success-based metrics share a common formulation:  
\begin{equation}
\text{Success}(M) = \frac{1}{N} \sum_{i=1}^{N} S_i^{(M)},
\end{equation}
where $N$ is the total  \#trials, $S_i^{(M)} \in \{0,1\}$ is a binary indicator of whether trial $i$ is successful under metric $M$. Different metrics correspond to different success conditions:
\begin{itemize}
    \item \textbf{Detection Success Rate (DSR):} $S_i^{(\text{DSR})}=1$ if the target object is correctly detected from the initial camera observation.  
    \item \textbf{Grasp Success Rate (GSR):} $S_i^{(\text{GSR})}=1$ if the robot successfully grasps and lifts the target object to a predefined height.  
    \item \textbf{Task Success Rate (TSR):} $S_i^{(\text{TSR})}=1$ if the grasp both succeeds and satisfies task-specific constraints.  
\end{itemize}

We further evaluate execution efficiency using the success weighted by path length (SPL) metric:
\begin{equation}
\text{SPL} = \frac{1}{N} \sum_{i=1}^{N} S_i \times \frac{l_i^*}{\max(l_i^*, l_i)},
\end{equation}
where $S_i$ is a binary success indicator, $l_i^*$ is the shortest path length, and $l_i$ is the actual executed path length.

To compare run-time efficiency across methods, we report the Execution Time (ET) in seconds for each approach. All experiments were conducted on the same hardware, an Intel Xeon 4215R (3.2 GHz) CPU and an NVIDIA GeForce RTX 4080 GPU, to ensure a consistent and fair comparison.
\section{Experiments} \label{Sec: experiments}
Since our goal is to benchmark the grasping effectiveness of recent methods, especially VLA models, we first collect a dataset to enable fine-tuning of these models. In our preliminary experiments, zero-shot use of these approaches shows significantly low accuracy on the complex tasks required by our design, so fine-tuning is necessary to obtain reliable results for a fair comparison. This is also widely observed by several recent works~\cite{han2024fetchbench,khazatsky2024droid}.

\begin{figure}[t]
    \centering
    \includegraphics[width=0.95\linewidth]{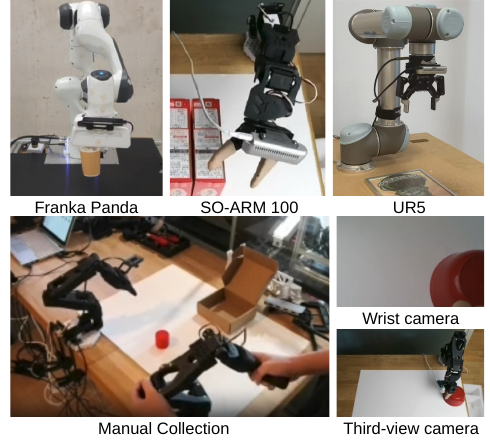}
    \vspace{1ex}
    \caption{Real-world data collection setup with three robotic arms: Franka Emika Panda, SO-ARM 100, and UR5. We employ one wrist-mounted camera and one third-view camera, capturing both RGB and depth images.}
    \label{fig:real-world_robot_data}
    \vspace{-1ex}
\end{figure}

\subsection{Data Collection}
\begin{figure*}[t]
    \centering
        \includegraphics[width=0.90\linewidth]{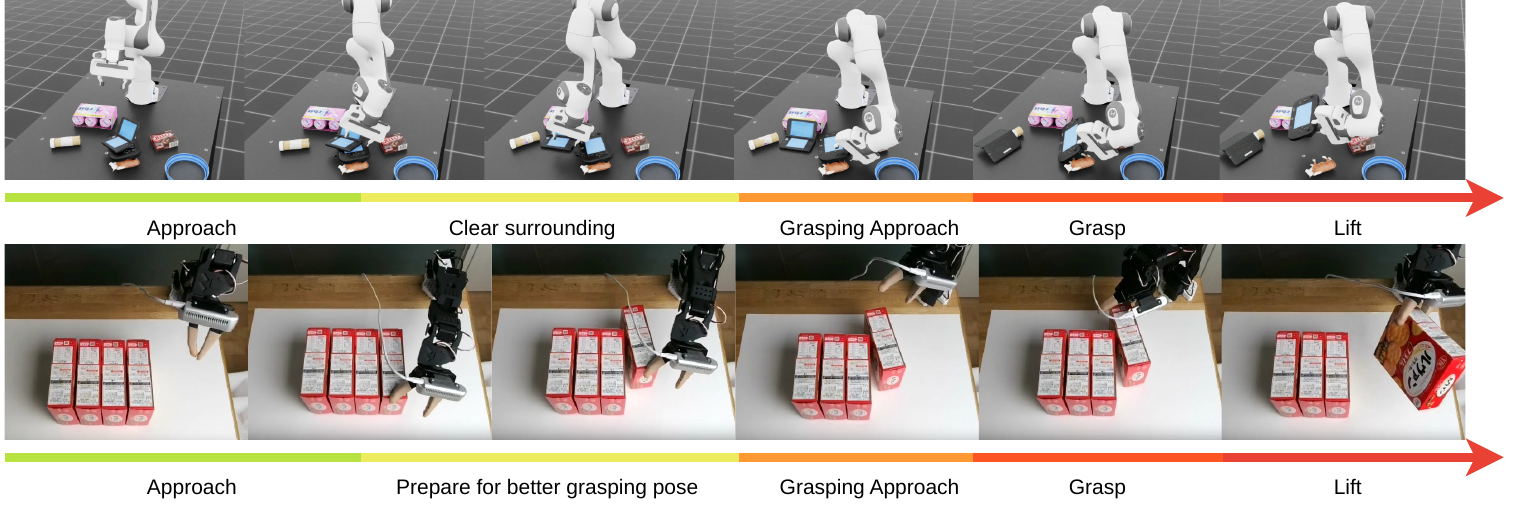}
    \vspace{2ex}
    \caption{Action complexity from our collected data. Complex cluttered scenes cannot be solved by visual detection alone and require sufficient trajectory planning. For example, in stacked object scenarios, surrounding objects need to be cleared to fetch the target safely, and additional actions such as pushing are required to prepare for a better grasping pose.}
    \label{fig:trajectory}
\end{figure*}

Since many of our tasks involve complex object interactions that remain unsolved by current autonomous grasping systems, we collect trajectories manually. For each task, we vary trajectories and grasp poses to ensure diversity and robustness. In simulation, we use a 3Dconnexion SpaceMouse, which allows smooth trajectory demonstrations with fine-grained control. For real-world experiments, we employ the robot’s teaching mode and SpaceMouse, enabling natural teleoperation and precise execution. We collect both successful and failed trials. Failure cases, such as drops, missed grasps, or collisions, provide informative negative supervision for learning-based methods and allow fine-grained analysis of task difficulty.   
Each trajectory includes synchronized observations from both a wrist-mounted and a third-person camera, recording RGB and depth images. In addition, we log robot state information, including end-effector pose, joint positions, and executed actions. The action space is represented as the delta of the end-effector position and orientation \([x, y, z, r, p, y]\) with the binary gripper state. To this end, we collected a dataset with $5000$ simulation trajectories and $800$ trajectories from real robots to support fine-tuning. Fig.~\ref{fig:real-world_robot_data} shows our real-world data collection setup, and Fig.~\ref{fig:trajectory} shows examples of collected trajectories.

\begin{table*}[t]
  \centering
  \caption{Comparison of baseline methods on GCA-Bench.}
  \vspace{1em}
  \scalebox{0.96}{
  \begin{tabular}{lcccccccccccc}
    \toprule
     & \multicolumn{3}{c}{Singulated}
     & \multicolumn{3}{c}{Clutter}
     & \multicolumn{3}{c}{Constrained Space}
     & \multicolumn{3}{c}{Semantic} \\
    \cmidrule(r{6pt}){2-4}
    \cmidrule(l{6pt}r{6pt}){5-7}
    \cmidrule(l{6pt}r{6pt}){8-10}
    \cmidrule(l{6pt}){11-13}
 Method & TSR$\uparrow$ & SPL$\uparrow$ & ET$\downarrow$ & TSR$\uparrow$ & SPL$\uparrow$ & ET$\downarrow$ & TSR$\uparrow$ & SPL$\uparrow$ & ET$\downarrow$ & TSR$\uparrow$ & SPL$\uparrow$& ET$\downarrow$ \\
    \midrule
    AnyGrasp~\cite{fang2023anygrasp}        &  0.37    & 0.15 &  13.07    & 0.27       & 0.11   &   \underline{13.07}      & 0.13 &0.19 & \textbf{9.05}  & --   & --   & --  \\
    GraspMAS~\cite{nguyen2025graspmas}        & 0.23 &   0.18   &  30.47   &0.07 & 0.07  & 46.14   & 0.14   & 0.28   & 31.88 &0.15 &0.73 &45.24  \\
    \midrule
    AnyGrasp~\cite{fang2023anygrasp} + cuRobo~\cite{sundaralingam2023curobo}   & 0.41     &  0.68    &  12.41    & 0.27 &   0.58   &  \textbf{12.74}   & 0.32    &   0.69  &  \underline{9.65}  & --   & --   & -- \\
    GraspMAS~\cite{nguyen2025graspmas} + cuRobo~\cite{sundaralingam2023curobo}  &   0.28   &  0.72    & 29.78     &  0.08&  0.67  &    43.72   &  0.11   &  \textbf{0.72}   &   30.46   & 0.16  & \textbf{0.84}  &45.02  \\
    \midrule
    GraspVLA~\cite{deng2025graspvla}       &  0.07    & 0.43     &  223.15    &  0.08 &  0.35    &  513.61    &  0.05   & 0.51    &  194.50   &  0.08   &  \underline{0.75}   &  153.84   \\
    OpenVLA~\cite{kim2024openvla}         &    0.58  &  0.54    &   97.64   &  0.42    &  0.68     &   203.60   &   0.41  &  0.65   & 139.04   &  0.43    &  0.68   &  174.17 \\
    OpenVLA-OFT~\cite{kim2025fine}          &    \underline{0.73}  &  0.68    &   \textbf{8.32}   &  \underline{0.57}    &  \underline{0.71}     &   24.51   &   \textbf{0.51}  &  0.52   & 15.6   &  0.42    &  0.70&  18.36\\
    $\pi_0$~\cite{black2024pi_0}         &    \textbf{0.77}  &  \textbf{0.77}    &   \underline{9.37}   &  0.52    &  0.64     &   22.65   &   \textbf{0.51}  &  \underline{0.71}   & 14.4   &  \underline{0.45}    &  0.61   &  \underline{17.7}\\
    $\pi_{0.5}$~\cite{intelligence2025pi_}         &    \textbf{0.77}  &  \underline{0.73}    &   13.35   &  \textbf{0.61}    &  \textbf{0.81}    &   16.77   &   \underline{0.46}  &  0.16   & 13.35   &  \textbf{0.53}    &  0.65   &  \textbf{14.85}\\
    
    \bottomrule
  \end{tabular}
  }
  
  \label{tab:baseline}
\end{table*}
\subsection{Baselines}
To evaluate the performance of existing grasping systems on GCA-Bench, we consider representative baselines from two major paradigms: (\textit{i}) grasp detection with motion planning and (\textit{ii}) end-to-end VLA policies.
For the grasp detection pipeline, we include AnyGrasp~\cite{fang2023anygrasp}, a state-of-the-art 6-DoF grasp detection model, and GraspMAS~\cite{nguyen2025graspmas}, a multi-agent framework that leverages large language models for language-conditioned grasp generation. To assess the impact of motion feasibility and collision avoidance, we evaluate these predicted grasps both independently and in combination with cuRobo~\cite{sundaralingam2023curobo}, which performs collision-aware trajectory optimization and control.
For VLA models, we first evaluate GraspVLA~\cite{deng2025graspvla} in a zero-shot setting, which leverages large dataset for strong zero-shot generalization. We further consider general VLA models finetuned on our dataset, including OpenVLA~\cite{kim2024openvla} and its variant OpenVLA-OFT~\cite{kim2025fine}, as well as models from the $\pi$-series: $\pi_0$~\cite{black2024pi_0} and $\pi_{0.5}$~\cite{intelligence2025pi_}. These models are fine-tuned using LoRA on collected dataset, for 60k steps on 4 NVIDIA A100 GPUs.

For each task category, we sample 5 scenarios. Within each scenario, we randomize object poses and arrangements across 10 trials, yielding 50 trials per task category. Based on the design of GCA-Bench, we conduct experiments to address the following research questions:

\textbf{Q1:} How do existing baselines perform on our benchmark? Can they support complex grasping tasks?

\textbf{Q2:} What is the gap between visual-based grasp detection and real-world grasp execution? Can current motion planning or end-to-end algorithms bridge this gap?

\textbf{Q3:} Can GCA-Bench improve the performance on real-world tasks?

\subsection{Results}
\textbf{Q1: How do existing baselines perform?} Table~\ref{tab:baseline} shows that the overall success rate across all four major task categories remains below 70\% for each baseline, indicating that existing methods struggle to support complex grasping tasks. Detection-based pipelines achieve low task success rates on complex grasping. Although incorporating motion planning slightly improves performance by reducing collisions, it does not address the fundamental limitations of grasp detection. In particular, traditional pipelines lack adaptive replanning and task-level reasoning, which are critical for multi-step or constraint-heavy grasping. Despite its claimed zero-shot generalization ability, GraspVLA fails on most tasks, likely because its training data is biased toward standard grasping settings.
Fine-tuned VLA models demonstrate substantially better performance, with $\pi_{0.5}$ achieving the best overall results. However, performance still degrades as task complexity increases. Moreover, the relatively low success rate on semantic tasks suggests that current VLA models still have limited high-level reasoning capabilities. 

Table~\ref{tab:scene_results} breaks down the performance of different methods across scenes of varying difficulty. This table shows that detection-based two-stage methods are efficient and robust in simple tasks such as grasping common objects and scattered scenes, but fail in complex scenarios due to the lack of real-time feedback. Adding collision-free motion planning improves performance in constrained environments but can sometimes reduce success, for example, in cluttered scenes, where incidental contact can facilitate stable grasps. These results show that current pipelines are brittle and far from robust across diverse complex tasks.  
\begin{table}[h]
\caption{Evaluation results across different task scenarios.}
\centering
\vspace{1em}
\setlength{\tabcolsep}{2pt}
\scalebox{0.83}{
\begin{tabular}{lcccccccc}
\toprule
 & \multicolumn{3}{c}{Singulated} & \multicolumn{3}{c}{Clutter} & \multicolumn{2}{c}{Constrained} \\
\cmidrule(lr){2-4} \cmidrule(lr){5-7} \cmidrule(lr){8-9}
Method & simple & complex & flat & scattered & stacked & packed & basket & shelf \\
\midrule
AnyGrasp~\cite{fang2023anygrasp}             & 0.62 & 0.50 & 0.0  & 0.44 & \textbf{0.26} & 0.12 & 0.22 & 0.04 \\
GraspMAS~\cite{nguyen2025graspmas}            & 0.32 & 0.36          & 0.0  & 0.08 & 0.12          & 0.02          & 0.28 & 0.0  \\
AnyGrasp~\cite{fang2023anygrasp}+cR~\cite{sundaralingam2023curobo} & 0.76 & 0.48 & 0.0  & 0.48 & 0.22 & 0.10 & 0.54 & 0.10 \\
GraspMAS~\cite{nguyen2025graspmas}+cR~\cite{sundaralingam2023curobo} & 0.44 & 0.40 & 0.0  & 0.08 & 0.12 & 0.04 & 0.22 & 0.0  \\
GraspVLA~\cite{deng2025graspvla}             & 0.18 & 0.02          & 0.0  & 0.14 & 0.06          & 0.04          & 0.10 & 0.0  \\
OpenVLA~\cite{kim2024openvla}  & 0.88  & 0.62     & 0.24   & 0.53      & 0.08      & 0.64   & 0.60    & 0.22          \\                                      
OpenVLA-OFT~\cite{kim2025fine}     & \textbf{0.98}  & \textbf{0.82}     & 0.40   & \underline{0.78}      & 0.20      & \underline{0.74}   & \underline{0.66}    & \underline{0.36}   \\
    $\pi_0$~\cite{black2024pi_0} &\underline{0.96}&0.74&\textbf{0.58}&0.72&0.12&0.72&0.50&0.34\\
    $\pi_{0.5}$~\cite{intelligence2025pi_}&\textbf{0.98}&\underline{0.76}&\underline{0.56}&\textbf{0.82}&\underline{0.24}&\textbf{0.76}&\textbf{0.68}&\textbf{0.42}\\                                                         
\bottomrule
\end{tabular}
}
\label{tab:scene_results}
\end{table}

Table~\ref{tab: semantic} further evaluates semantic instruction-following performance under increasing levels of complexity. GraspMAS maintains reasonable performance on basic and simple-constraint tasks by leveraging LLM-based reasoning, but its reasoning is confined to localizing the object or part specified by the instruction and generating a grasp pose on it. In contrast, VLA models rely on action patterns learned from demonstrations rather than explicit language understanding. They perform well when instructions closely match the training distribution but struggle with novel semantic constraints, revealing that their instruction-following capability is largely a form of pattern matching rather than true task reasoning. 
\begin{table}[h]
\caption{Instruction following results}
\centering
\vspace{1em}
\scalebox{0.85}{
\begin{tabular}{lcccccccccccc}
\toprule
 & \multicolumn{2}{c}{Basic} & \multicolumn{2}{c}{Simple Constraint} & \multicolumn{2}{c}{High-level} \\
\cmidrule(r{4pt}){2-3} \cmidrule(l{4pt}r{4pt}){4-5} \cmidrule(l{4pt}){6-7}
Method & GSR$\uparrow$ & TSR$\uparrow$  & GSR$\uparrow$ & TSR$\uparrow$ & GSR$\uparrow$ & TSR$\uparrow$  \\
\toprule
GraspMAS~\cite{nguyen2025graspmas}     & 0.46   &  0.42    & 0.42    &   0.16  & 0.12&  0.04       \\
GraspMAS~\cite{nguyen2025graspmas}+cR~\cite{sundaralingam2023curobo}    & 0.46   &  0.42    & 0.44     &  0.16   &  0.12  &  0.04         \\
GraspVLA ~\cite{deng2025graspvla}         &   0.2   &   0.16   &  0.12   &  0.08      &  0.06 & 0.0       \\
OpenVLA~\cite{kim2024openvla}      & 0.76      &0.76 &  0.78 & \underline{0.32}  &  0.68  &0.20   \\
OpenVLA-OFT~\cite{kim2025fine}      & 0.82   & 0.80&0.80 & 0.30  &  \underline{0.82}&0.16  \\
    $\pi_0$~\cite{black2024pi_0} &\underline{0.86} &\underline{0.82} &\underline{0.84} & 0.28 & 0.78&\underline{0.24}\\
    $\pi_{0.5}$~\cite{intelligence2025pi_}&\textbf{0.92} &\textbf{0.92} &\textbf{0.90}&\textbf{0.42}&\textbf{0.88}&\textbf{0.30}\\

    \bottomrule
\end{tabular}
}
\label{tab: semantic}
\end{table}

\textbf{Q2: What is the gap between detection and execution?} Table~\ref{tab:gsr_dsr_ratio} reveals the gap between detection capability and execution success by normalizing grasp success rates against detection performance. The results show that the ratio is consistently below 0.5, indicating that even when objects are successfully detected, executing stable grasps in complex scenarios remains challenging. To analyze this gap more systematically, we break down the failures into three levels: detection, grasp, and task execution, each revealing different limitations of the pipeline. (\textit{i}) Detection Failures: Common objects are reliably detected, whereas irregular shapes and occluded items remain challenging. Robustness could be improved through multi-view or temporal fusion, and further enhanced with self-supervised refinement or online adaptation in cluttered or partially observed environments. \textit{(ii)} Grasp Failures: Small, thin, or slippery objects frequently slip during execution, even when a valid grasp pose is detected. Cluttered and constrained spaces further exacerbate failures by restricting maneuverability. These cases often require preparatory actions or adaptive strategies. \textit{(iii)} Task Failures: As shown in Table~\ref{tab: semantic}, for tasks with semantic or physical constraints, such as grasping a filled cup, open-loop pipelines fail due to the absence of feedback. We observe that current methods cannot detect or compensate when objects tilt or when collisions occur, leading to frequent task failures. This underscores the need for closed-loop control and adaptive planning in complex manipulation tasks. 
\begin{table}[h]
\caption{Detection-normalized grasp success rates.}
\centering
\vspace{1em}
\scalebox{0.98}{
\begin{tabular}{lccc}
\toprule
Method & Singulated & Clutter & Constrained \\
\midrule
AnyGrasp~\cite{fang2023anygrasp}     & \underline{0.37} & \underline{0.27} & 0.14 \\
GraspMAS~\cite{nguyen2025graspmas}     & 0.28 & 0.12 & \underline{0.23} \\
AnyGrasp~\cite{fang2023anygrasp} + cuRobo~\cite{sundaralingam2023curobo}& \textbf{0.41} & \textbf{0.28} & \textbf{0.34} \\
GraspMAS~\cite{nguyen2025graspmas} + cuRobo~\cite{sundaralingam2023curobo}& 0.34 & 0.12 & 0.18 \\
\bottomrule
\end{tabular}
}
\label{tab:gsr_dsr_ratio}
\end{table}

\textbf{Q3: Can GCA-Bench improve the performance on real-world tasks?} 
\begin{figure}
    \centering
    \includegraphics[width=0.8\linewidth]{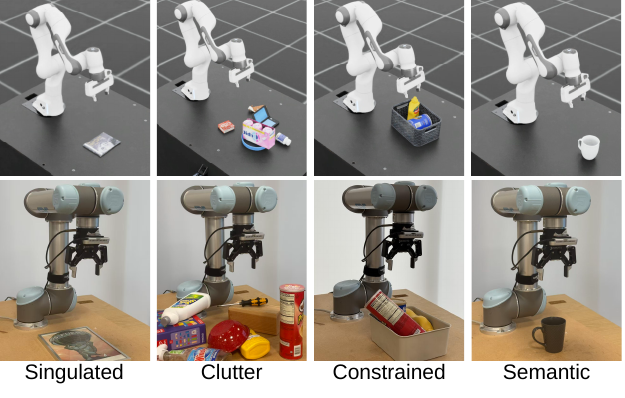}
    \vspace{2ex}
    \caption{Paired task scenarios for real-world validation.}
    \label{fig:experiment setup}
    \vspace{-1ex}
\end{figure}
\begin{figure}
    \centering
    \includegraphics[width=0.8\linewidth]{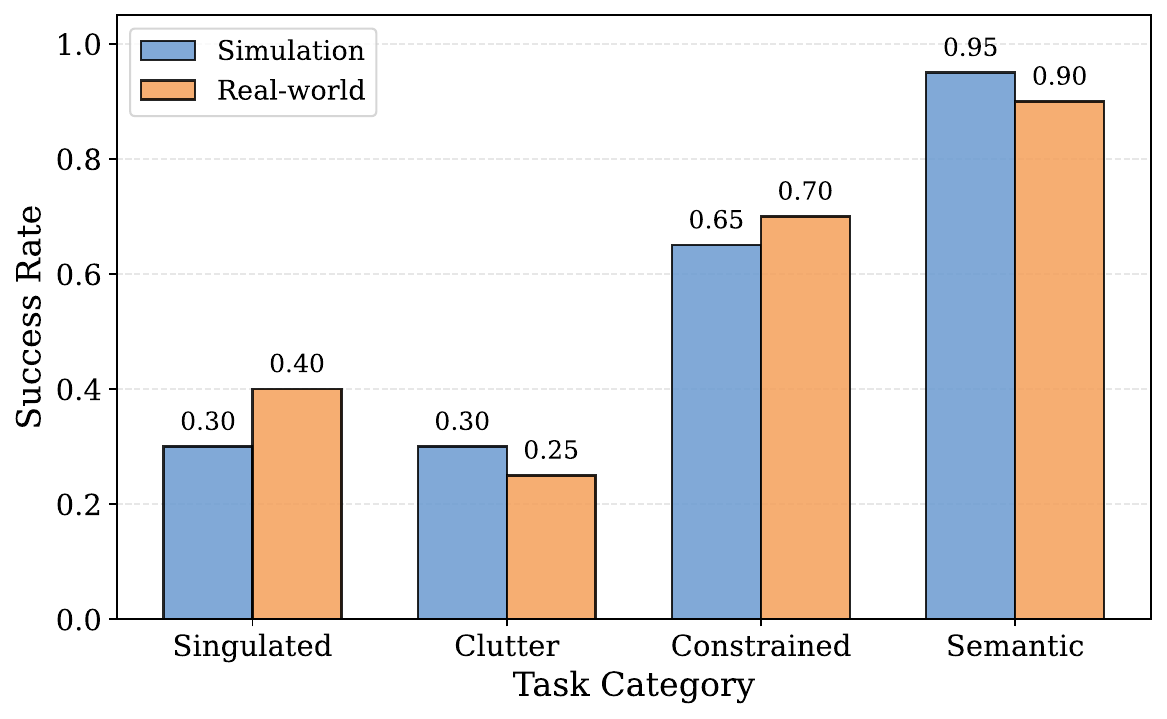}
    \vspace{2ex}
    \caption{Real-world performance with simulation baseline.}
    \label{fig:realworld_success}
        \vspace{-1ex}
\end{figure}
To validate that GCA-Bench provides a meaningful evaluation platform and supports deployment on real-world grasping tasks, we conduct experiments using a UR5 robot with a Robotiq 2F-85 gripper. We evaluate four representative tasks for each category as shown in Fig.~\ref{fig:experiment setup}. For each task, the fine-tuned $\pi_{0.5}$ policy is tested over 20 trials and compared with simulation results on our benchmark. As shown in Fig.~\ref{fig:realworld_success}, the results reveal consistent performance. Singulated and Clutter remain challenging due to precise multi-step manipulation requirements, while Constrained tasks benefit from collision-aware behaviors learned in simulation. Semantic tasks achieve the highest success rate, indicating that diverse language instructions in our dataset improve instruction-following in real-world settings. Notably, real-world performance closely aligns with simulation results, suggesting that GCA-Bench provides a reliable and practical benchmark suite for developing grasping methods capable of handling complex real-world grasping.

\begin{figure}
    \centering
    \includegraphics[width=0.9\linewidth]{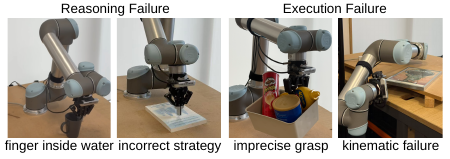}
    \vspace{2ex}
    \caption{Failure cases.}
    \label{fig:failure}
        \vspace{-1ex}
\end{figure}

Fig.~\ref{fig:failure} shows representative failure cases. For unseen instruction, the policy struggles with high-level semantic reasoning and object affordance understanding (e.g., “pass me a cup of water”). Similarly, for unseen flat objects, the policy fails to associate them with similar tasks present in the training dataset. For seen tasks, failures mainly arise from imprecise grasps, leading to incorrect object misgrasp or unstable grasps. Additionally, the VLA policy occasionally predicts actions that lead to kinematically infeasible configurations without recovery, highlighting a gap between action prediction and low-level motion feasibility.

\section{Conclusions}\label{Sec: conclusion}
We introduce GCA-Bench, a benchmark for evaluating robotic grasping across the full pipeline under both scene-level and semantic complexity. Our experiments show that existing methods perform well in simple settings, but success rates drop sharply in cluttered, constrained, or instruction-driven tasks, exposing a persistent gap between perception and reliable execution. GCA-Bench is additionally well-suited for testing VLA models because it couples language-conditioned, semantically grounded grasp goals with the full demands of execution, enabling a direct assessment of whether VLAs truly ground instructions into robust actions rather than relying on perception or unconstrained motions. A current limitation is that GCA-Bench focuses on parallel-jaw grippers; future work will expand to more diverse gripper types and settings. GCA-Bench will be made publicly available to support the benchmarking and future study.

\bibliographystyle{class/IEEEtran}
\bibliography{class/IEEEabrv,class/reference}
   
\end{document}